\pdfoutput=1

\documentclass[11pt]{article}

\usepackage{EMNLP2023}

\usepackage{times}
\usepackage{latexsym}
\usepackage{graphicx}
\usepackage{comment}

\usepackage[T1]{fontenc}

\usepackage[utf8]{inputenc}

\usepackage{microtype}

\usepackage{inconsolata}
\usepackage{xurl}

\title{Regulation and NLP (RegNLP): Taming Large Language Models}

\author{Catalina Goanta \\
  Utrecht University \\ \And
  Nikolaos Aletras \\
  University of Sheffield \\\And
  Ilias Chalkidis \\
  University of Copenhagen \\\AND
  Sofia Ranchordas \\
  Tilburg University \\\And
  Jerry Spanakis \\
  Maastricht University}

\begin{document}
\maketitle
\begin{abstract}
The scientific innovation in Natural Language Processing (NLP) and more broadly in artificial intelligence (AI) is at its fastest pace to date. As large language models (LLMs) unleash a new era of automation, important debates emerge regarding the benefits and risks of their development, deployment and use.
Currently, these debates have been dominated by often polarized narratives mainly led by the \emph{AI Safety} and \emph{AI Ethics} movements. This polarization, often amplified by social media, is swaying political agendas on AI regulation and governance and posing issues of regulatory capture. Capture occurs when the regulator advances the interests of the industry it is supposed to regulate, or of special interest groups rather than pursuing the general public interest. Meanwhile in NLP research, attention has been increasingly paid to the discussion of regulating risks and harms. This often happens without systematic methodologies or sufficient rooting in the disciplines that inspire an extended scope of NLP research, jeopardizing the scientific integrity of these endeavors. \textit{Regulation studies} are a rich source of knowledge on how to systematically deal with \textit{risk and uncertainty}, as well as with \textit{scientific evidence}, to evaluate and compare regulatory options. This resource has largely remained untapped so far. In this paper, we argue how NLP research on these topics can benefit from proximity to regulatory studies and adjacent fields. We do so by discussing basic tenets of regulation, and risk and uncertainty, and by highlighting the shortcomings of current NLP discussions dealing with risk assessment. Finally, we advocate for the development of a new multidisciplinary research space on regulation and NLP (RegNLP), focused on connecting scientific knowledge to regulatory processes based on systematic methodologies.

\end{abstract}


\section{Introduction}

The development of Large Language Models (LLMs) is at its fastest pace to date. In the past years alone, LLMs have seen considerable advancement across a multitude of languages and types of data, with models such as GPT-3.5~\cite{instructgpt}, GPT-4~\cite{openai2023gpt4}, LLaMA~\cite{touvron2023llama}, and PALM-2~\cite{anil2023palm} demonstrating unprecedented capabilities across a broad collection of natural language processing (NLP) tasks.\footnote{In the sense that LLMs generalize to a great extend in out-of-distribution and out-of-domain use cases.} 

\begin{figure}
    \centering
    \resizebox{\columnwidth}{!}{
    \includegraphics{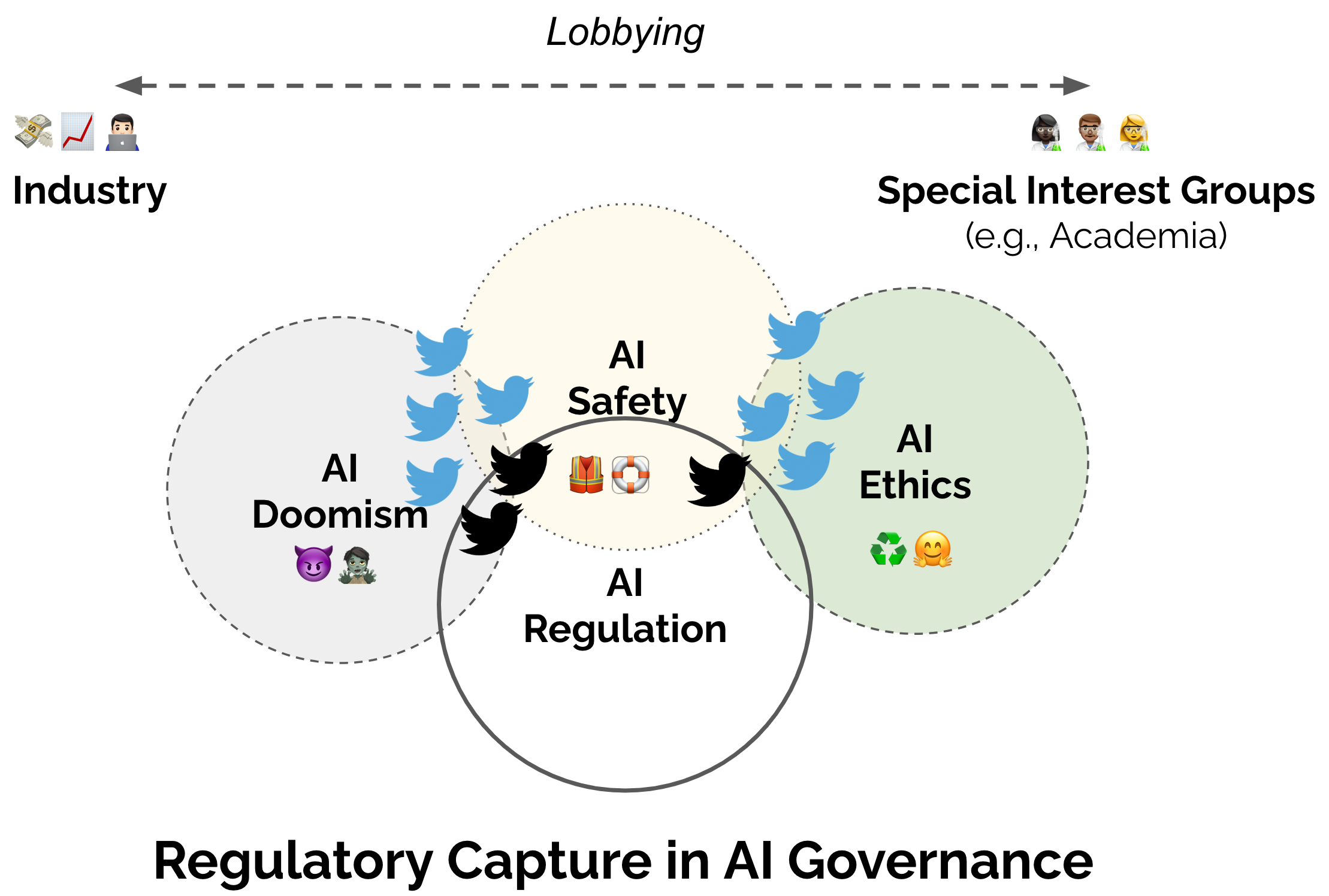}
    }
    \caption{A depiction of the cross-over between AI Safety, Ethics, Doomism, and how they capture AI Regulation.}
    \label{fig:enter-label}
    \vspace{-4mm}
\end{figure}

These innovations have led to rapid shifts in various applications such as open-domain search, coding, e-commerce and education. For example, state-of-the-art LLMs already power conversational search engines (e.g. OpenAI ChatGPT, Bing Chat, and Google Bard), coding assistants (e.g. OpenAI Codex and Github Copilot), product recommender systems (e.g. Alibaba Tongyi and SalesForce CommerceGPT) and educational assistants (e.g. Khanmigo) inter alia. 

As with most technologies, the development and use of LLMs do not come without concerns. Researchers are rightfully worried that while this technology may be transformative, its societal implications might be higher than its benefits \cite{gabriel-2020-alignment}. Concerns have been raised especially around ethics \cite{Floridi2023-vl, tsarapatsanis-aletras-2021-ethical}, bias \cite{Hovy2021-wt, blodgett-etal-2020-language}, safety \cite{Dobbe2021-wa} and environmental impact \citep{Rillig2023-kp, Schwartz2020-df, strubell-etal-2019-energy}. The unsupervised use of LLMs has already led to widely-publicized examples of professional negligence. The all too public fiasco of the lawyer who used ChatGPT for a court brief and unknowingly included made-up case law references was taken by many as an example of the dangers of immersing daily professional activities in generative AI.\footnote{\url{https://edition.cnn.com/2023/05/27/business/chat-gpt-avianca-mata-lawyers/index.html}} The public debate around the future of AI, and implicitly NLP, is very complex and multi-layered. While the debate seems to converge on the point of calling for regulation to control the unwanted effects of these technologies,\footnote{\url{https://www.forbrukerradet.no/side/new-report-generative-ai-threatens-consumer-rights/}} different regulatory directions are proposed by the various stakeholders involved in this debate.   


The current public discourse has been dominated by two groups. 
On the one hand, proponents of AI existential risks (the \emph{AI Safety} movement)\footnote{We distinguish the AI Safety movement from `AI Doomism' (\url{https://time.com/6266923/ai-eliezer-yudkowsky-open-letter-not-enough/}) flirting with conspiracy theories.} that include technology CEOs and AI researchers have been publishing open letters\footnote{\url{https://www.nytimes.com/2023/03/29/technology/ai-artificial-intelligence-musk-risks.html}}\textsuperscript{,}\footnote{\url{https://futureoflife.org/open-letter/pause-giant-ai-experiments/}}\textsuperscript{,}\footnote{\url{https://www.safe.ai/statement-on-ai-risk}} 
and regularly meet with regulators to warn about catastrophic scenarios around general AI, proposing industry-friendly solutions \cite{nytimesAIPoses}. On the other hand, the \emph{AI ethics} movement \citep{Borenstein2021-ws}, mostly reflects the voice of researchers from various disciplines as well as civil society activists. They have raised compelling alarm bells with respect to the risks posed by LLMs \citep{Bender2021, weidinger, Floridi2023-vl}. The AI ethics movement has also offered guidance to regulators such as the European Parliament on AI governance, arguing for e.g. broad definitions of general purpose AI.\footnote{\url{https://ainowinstitute.org/publication/gpai-is-high-risk-should-not-be-excluded-from-eu-ai-act.}}




Yet while NLP research increasingly focuses on LLM regulation, it remains generally detached from prior work on regulation studies. Instead, the often intense public conflicts in this space are nudging regulators towards reactionary public relations activities rather than the collection of scientific expertise representing broader parts of the NLP and AI communities. For instance, right after the most recent letter on existential risks, the US and EU agreed to develop an AI code of conduct `within weeks'.\footnote{\url{https://www.fastcompany.com/90903919/will-the-eu-u-s-new-voluntary-code-of-conduct-on-ai-work-to-rein-in-the-tech}} Similarly, the UK has announced it will be holding a global summit on AI safety,\footnote{\url{https://www.theguardian.com/technology/2023/jun/09/rishi-sunak-ai-summit-what-is-its-aim-and-is-it-really-necessary.}} and the US Congress has been taking evidence from a wide array of industry executives, in what has been described as a global race to regulate AI. \footnote{\url{https://foreignpolicy.com/2023/05/05/eu-ai-act-us-china-regulation-artificial-intelligence-chatgpt/}} With legacy and social media considerably increasing the visibility of these often polarizing debates, there is a real danger of regulatory capture by visible voices in industry and academia alike on scientific views that might not necessarily be representative of the `silent majority' of NLP researchers. Regulatory capture is the process by which regulation is directed away from the public interest and towards the interests of specific groups \cite{levine, Dal_Bo2006-hn}. 

Against this background, we call on the NLP community to familiarize itself with regulation studies. We argue that this can lead to a clearer vision about how NLP as a field can properly participate in AI governance not only as an object of regulation, but also as a source of scientific knowledge that can benefit individuals, societies and markets alike. In this paper, we contribute to the existing debate relating to the future of NLP by discussing the benefits of interfacing NLP research with regulation studies in a systematic way. This view is based on two main ideas: 

\begin{enumerate}
\item  NLP research on regulation needs a multidisciplinary framework engaging with regulation studies, as well as adjacent disciplines such as law, economics, environmental science, etc. We advocate for a new crucial area of research on regulation and NLP (RegNLP), with harmonized and systematic methodologies.  



\item  A more coordinated NLP research field on risk and regulation (such as RegNLP) can interact with policy-makers with more transparency, representation, and trustworthiness. 

\end{enumerate}


\section{Regulation: A Short Introduction}

\paragraph{Why Do We Regulate?}

Calls to regulate  LLMs and AI are everywhere, to the extent that overusing the term `regulation' is trivializing its meaning. So what exactly \textit{is} regulation and why do we rely on it?

Historically, regulation was defined by reference to state intervention in the economy. \citet{selznick1985focusing} defined regulation as `a sustained and focused control exercised by a public agency over activities that are valued by the community'. Over the last decades, regulation has evolved and it has increasingly acquired a hybrid character as both public and private actors may issue rules that shape social behavior. In this paper, we draw on Black's definition of regulation as an organized and intentional attempt to manage another person's behavior so as to solve a collective problem \cite{Black2008-uk}. This is done through a combination of rules or norms which come together with means for their implementation and enforcement \cite{Ogus}.

Regulation includes different types of regulatory instruments (e.g. laws) such as traditional top-down or command-and-control regulations. Recent examples in digital public policy include the laws issued by the European Union, such as the Digital Services Act, or India's law banning TikTok. However, beyond these public regulatory instruments, there is an array of private or hybrid instruments \cite{veale2023ai}. Some of these are qualified as `soft' regulation because they cannot be enforced in court but they remain relevant and they effectively shape the behavior of the industry. Examples include the EU's Ethics Guidelines for Trustworthy AI\footnote{\url{https://digital-strategy.ec.europa.eu/en/library/ethics-guidelines-trustworthy-ai}}, but also codes of conduct initiated by industry itself.\footnote{\url{https://ethics.acm.org/code-of-ethics/software-engineering-code/}}  

Over the last decades, different theories of regulation have helped us understand either on normative or empirical accounts why we should regulate. They generally reflect various hypotheses about `why regulation emerges, which actors contribute to that emergence and typical patterns of interaction between regulatory actors' \cite{Morgan2007-he}. Public interest and private interest theories are two well-known examples \cite{Morgan2007-he}. According to public interest theories, regulation is used by law-makers to serve a broad public interest, seeking to regulate in the most efficient way possible. Regulators assume that markets need `a helping hand' because when unhindered, they will fail. Information asymmetries are one of the market failures that regulation seeks to address. At the same time, public interest theories of regulation are normative and prescriptive: on the one hand, they assume that benevolent state regulators ought always to use regulation to advance the public interest; on the other, they also advise on how to achieve this goal. 

In contrast, private interest theories are not primarily concerned with normative justifications for regulation. Rather, they are prescriptive accounts of the complex dynamics between different market actors, stakeholders and public officials situated in a given socio-economic, political and cultural time and place. They explain, for example, why regulation may fail to pursue the public interest but they do not offer prescriptions on how address to such problems. Private interest theories assume that regulation emerges from the actions of individuals or groups motivated to maximize their self-interest. 


\paragraph{Technology Regulation and the Role of Science}

Technological change often disrupts the wider regulatory order, triggering concerns about its adequacy and regulatory legitimacy \cite{Brownsword2017-gq}. Differences in the timing of technology and regulation explain this difficulty. The literature has claimed there is a `pacing gap' between the slow-going nature of regulation and the speed of technological change \cite{marchant}. Technological innovations have specific development trajectories, investment and life cycles, and path dependencies \cite{Van_den_Hoven2014-xn} that do not go well with the speed of technology. This also applies to LLMs. This is a well-known problem in regulatory studies that has been captured by the Collingridge dilemma \cite{Genus2018-uq}. This dilemma explains that when an innovation emerges, regulators hesitate to regulate due to the limited availability of information. However, by the time more is known, regulations may have become obsolete as technology may have already changed. 


The increased pace of regulatory activities in the past years in the field of technology shows that regulators are trying to close the pacing gap and be more proactive in tackling the potential risks of technology. In doing so, they increasingly depend on retrieving scientific information in a quick and agile way. 

The role of science in regulation and public policy has been the subject of important debates. On the one hand, regulation should reflect the latest scientific evidence and be evidence-based. An illustration of this approach is the European Union's `Better Regulation' agenda, a public policy strategy aiming to ensure that European regulation is based on scientific evidence, as well as the involvement of a wide range of stakeholders in the decision-making process \cite{Commission2023,simonelli_iacob_2021}. Citizens, businesses and any other stakeholders can submit their contributions to the calls for evidence, feedback and public consultations. For instance, 303 contributions were received on the AI Act proposal during 26 April 2021 and 6 August 2021, out of which 28\% from businesses, 24\% from business associations, 17\% from NGOs, and 6\% from academic/research institutions \cite{Commission2022}. On the other hand, science is complex and difficult to translate into regulatory measures. Science has thus been used to oversimplify regulatory problems and justify poor regulatory decisions based on the existence of scientific evidence pointing in a specific direction \cite{Porter2020-zx}. This is a particular danger in the LLM public debate. With science becoming increasingly complex, so do the scientific perspectives on how to proceed with this technology.  

\paragraph{Lessons to be Learned} The theoretical underpinnings of regulation have helped shape a cohesive understanding of the rationale behind regulatory activity. The interest in technology regulation, particularly for disruptive innovations such as LLMs, has exploded in past years across a wide array of scientific disciplines. While understandable, such popularity often leads to inquiries which are not connected to prior knowledge on regulation studies. This reflects a more general problem faced by contemporary science, namely that of tackling multidisciplinary issues without multidisciplinary expertise. Considering these circumstances, research on LLMs and regulation could benefit from engaging with the regulation studies and governance context.  


\section{LLMs: Risk and Uncertainty}


The need to bridge NLP research with regulation studies is especially important in the discussion of risks. New and emerging technologies are typically accompanied by risk and uncertainty. The regulation of technological change and innovation is highly complex, as innovation remains an elusive concept hard to define, measure, and thus regulate. 

In the past years, the question of risks arising out of NLP developments such as LLMs has been increasingly embraced in computer science literature. One strand of this literature is reflected by the theme of algorithmic unfairness. This theme emerged at the intersection of discrimination law and automated decision-making, and includes questions relating to fairness and machine learning in general \cite{barocas-hardt-narayanan}, as well as specific examples of algorithmic bias risks in NLP \cite{Field2023-de, talat-etal-2022-reap, Kidd2023-iw}, computer vision \cite{Wolfe}, multimodal models \cite{birhane2021multimodal}, as well as privacy risks \cite{mireshghallah-etal-2022-quantifying}. Another strand of this literature looks at LLMs from a more holistic perspective, raising concerns about their size vis-a-vis a broader number of risks for e.g. the environment, bias, representation or hate speech \cite{Bender2021, weidinger, bommasani2022opportunities}. This theme does not only include risks in commercial applications, but also risks arising out of the mere scientific development of technology.  

This literature has raised important concerns relating to the immediate and longer-term implications around the advancement of machine learning and NLP. However, when positioned in the regulatory context, we can observe conceptual clashes with frameworks which have been traditionally relied upon in public policy and risk regulation. One such framework is the field of risk and uncertainty. Put simply, `risk is the situation under which the decision outcomes and their probabilities of occurrences are known to the decision-maker, and uncertainty is the situation under which such information is not available to the decision-maker' \cite{Park2017-og}. In more technical terms, `risk is the probability of an event multiplied by its impact, and uncertainty reflects the accuracy with which a risk can be assessed' \cite{Krebs}. As a field, risk and uncertainty has made considerable contributions to the development of risk regulation, most notably in relation to environmental regulation \cite{Heyvaert2011-bf, Yarnold2022-ue}. It is important that policy-makers have a concrete quantification of risk \cite{Aumann2008-lk}, in order to determine the adequate level of risk associated with various public policies. In addition, risk determination and management have important economic consequences, specifically for determining `what level of expenditure in reducing risk is proportionate to the risk itself' \cite{Krebs}.  

Especially in the case of technologies that easily transcend physical borders, societies and economies, determining risk and uncertainty is a complex undertaking, even for scientists. Some of the factors that make it difficult to assess risk and uncertainty include the complexity of the technology itself, as well as the information asymmetry underlying commercial practices. LLMs are humongous (billion-parameters-sized) Transformer-based~\cite{Vaswani2017} models, which have been initially pre-trained as standard language models~\cite{radford2018improving} in a vast quantity of text data (mostly web scrapes) and have been also further optimized to follow instructions~\cite{chung2022scaling} and user \emph{alignment}~\cite{leike-etal-2018-alignnment} with reinforcement learning from human feedback (RLHF) \cite{christiano-etal-2017-drl,Stiennon2020}. Particularly, AI alignment has been a controversial topic, since it implies a broad consensus on what sort of values (standards) AI should align with~\cite{gabriel-2020-alignment}. As such, LLMs are complex technologies where in addition to risk, we also deal with considerable uncertainty, which can be, among others, descriptive (e.g. relating to the variables defining a system), or related to  measurement (e.g. uncertainty about the value of the variables in a system) \cite{Gough}. 

\paragraph{LLMs as the new GMOs?} Yet LLMs are neither the first nor the last technology development posing concerns about wide-spread risks. As an illustration, in the 1970s and 1980s, In-Vitro Fertilization (IVF) was a demonized scientific development considered inhumane, which led to nothing short of a large-scale moral panic \cite{Garber}. In the 2000s, concerns around Genetically Modified Organisms (GMOs) dominated media coverage in European and North American countries, in what was deemed a `superstorm' of moral panic and new risk discourses \cite{Howarth2013-rj}. These are only two examples of risk narratives that were amplified by media coverage in ways that overshadowed important scientific expertise. Yet through regulation supporting scientific advancement, their use today has become mundane as part of solving considerable society problems such as infertility or food availability. These comparisons by no means imply that earlier biotechnological innovations pose the same levels of risk as LLMs or should entail the same level of regulation. However, it is important to learn from our past experiences with technology how to distinguish between moral panics and real problems that need scientific solutions.   

Researchers are starting to develop concrete methodologies for the auditing of LLMs and related NLP technologies \cite{derczynski2023assessing},\footnote{\url{https://github.com/leondz/garak/}} as well as dealing with particular risks such as environmental impact \cite{rolnick}. These contributions are much needed, as they can be translated into concrete measurements of risk and uncertainty, and further lead to the development of policy options in risk management. However, these initiatives are so far too few, as no cohesive scientific approach exists on the assessment of the risk and uncertainty posed by LLMs. To date, even the most comprehensive overviews of LLM risks \cite{weidinger} lack basic methodological practices such as the systematic retrieval of information from the disciplines of inquiry \cite{Page2021-kx}. In some cases, strong projections about risk impact are made without any scientific rigor whatsoever \cite{hendrycks2023overview}. Similarly, \citet{Dobbe2021-wa} note that while many technical approaches, including approaches related to 'mathematical criteria for ``safety'' or ``fairness'' have started to emerge, `their systematic organization and prioritization remains unclear and contested.

As a result, the lack of systematization and methodological integrity in scientific work around LLM risks contributes to a credibility crisis which may impact regulation and governance directly. 

\paragraph{Innovation Governance and Risk Relativization}
Learning from earlier experiences with risk and uncertainty can also help NLP researchers understand how risk has been dealt with in other policy areas. One of the reasons why it is important to contextualize NLP research on risks into a broader regulatory landscape is because this area has already generated meaningful frameworks for the understanding of risk in the context of innovation governance. Such frameworks include for instance the principle of responsible innovation, which calls for `taking care of the future through collective stewardship of science and innovation in the present' \cite{Von_Schomberg2013-hx}. This is only one of the many other approaches that can guide decision-making on technology regulation \cite{Hemphill2020-gt}.

A regulatory angle can also help with the relativization of risk - that is, putting risks into perspective by considering other policy areas as well. For instance, the UK Risk Register 2020 discusses potential risks and challenges that could cause significant disruption to the UK \footnote{\url{https://www.gov.uk/government/publications/national-risk-register-2020}}. The report thematically groups six risks (malicious attacks, serious and organized crime, environmental hazards, human and animal health, major accidents and societal risks). This insight is useful in understanding the scale and diversity of risks that public policy needs to account for. Such awareness could also contribute to the generation of policy options that can put LLM risks into perspective in relation to other categories of risks like the ones mentioned above. Here, a more holistic perspective on risk could also take a sectoral approach to LLM risks. For instance, going back to the example of the lawyer who invented case law using ChatGPT, existing legal and self-regulatory frameworks already address the risk of negligence in conducting professional legal activities. Considering this context can contribute with insights into whether it is really necessary to treat LLM-mediated information as a novel danger. While technology has generated a broad digital transformation \cite{Verhoef2021-cj}, it adds layers to existing problems (e.g. social inequality) which need policy interventions independent from their digital amplification.  




\section{Scientific Expertise, Social Media and Regulatory Capture}
\label{sec:sci_expert_soc_media}

\paragraph{Regulatory Capture} As a source of evidence for policy-makers, scientific expertise has increasingly played a central role in regulation \cite{Paschke2019-nr}. In some supranational governance contexts, scientific expertise is called upon in procedures that often require a certain level of transparency. This is the case of the call for public comments which we have discussed above as part of the EU's `Better Regulation' agenda. However, in the past decade, the rise of science communication on social media has somewhat changed the interaction between policy-makers and scientists \cite{Van_Dijck2020-rm}. A lot of the public debate between stakeholders from industry, academia and policy relating to LLM risks is had on social media platforms such as Twitter. This can pose a regulatory capture problem. Regulatory capture occurs when the regulator advances the interests of the industry it is supposed to regulate, or of special interest groups rather than pursuing the public interest \cite{Carpenter2013-bv}. Regulatory capture fits within the private interest theories we explored in Section 2, and often refers to the influence exercised by industry over regulation processes \cite{Saltelli2022}. The most recent example is OpenAI's white paper suggesting narrow regulatory interpretations for general purpose high-risk AI systems to European regulators \footnote{\url{https://time.com/6288245/openai-eu-lobbying-ai-act/}}. Similarly, multiple technology executives of large companies using LLMs, such as HuggingFace or OpenAI have been testifying before the US Congress to propose industry-friendly interpretations of AI risks. This is happening in a context of existing concerns around the industry orchestration of research agendas in NLP~\cite{abdalla2023elephant}, and science~\cite{abdalla2021hoodie} in general. However, regulation can also be captured by special interest groups from civil society, and increasingly, academia. In this meaning, regulatory capture has a cultural or value-driven dimension that encompasses `intellectual, ideological, or political forms of dominance' \cite{Saltelli2022}. In a landscape where special interest groups are increasingly represented by popular science communicators,  
a lot of questions arise in relation to the power exercised by the rising impact of science influencers, whether from academic, journalism and industry environments \cite{zhang}. 

\paragraph{The Rise of Science Influencers} Traditionally, science communication has followed a conventional model dominated by professional actors gatekeeping information (e.g. scientists, journalists and government). Social media has led to the creation of a networked model of science communication, with underlying socio-technical and political power shifts \cite{Van_Dijck2020-rm}. Science influencers are rooted in this development, as well as the broader rise of social media influencers \cite{Goanta_Ranchordás_2020}. They also bring with them additional complexities. They may emerge from a scientific background, but may use their platforms both for professional as well as personal self-disclosure \cite{Kim2016-ie}. In doing so, they also become political influencers who 'harness their digital clout to promote political causes and social issues' \cite{Riedl2023-zc}. To signal the social media influence of such public opinion leaders, some media outlets even rank them for awards purposes,\footnote{\url{https://www.euractiv.com/section/digital/news/meet-the-2019-euinfluencer-awardees/}} or profile them vis-a-vis state of the art scientific expertise.\footnote{\url{https://spectrum.ieee.org/artificial-general-intelligence}}   

This development is especially important since some social media platforms are more relevant for regulators than others. A recent report published by Oxford University's Reuters Institute shows that Twitter is the platform politicians pay most attention to across all studied markets.\footnote{\url{https://reutersinstitute.politics.ox.ac.uk/digital-news-report/2023}} With this in mind, the polarized narratives around LLM risks unfolding on social media pose the danger that scientific expertise is only partially represented in public debates, in spite of the promises of speech democratization expected from these platforms in the context of science communication. 

While followers and engagement may be a measure of popularity with some communities and stakeholders, it raises concerns relating to the role popularity metrics and algorithmic amplification on social media may have in representing scientific or industry consensus before policy-makers. As \citet{Zhang2018-nk} put it, there is a popularity bias that means `attention tends to beget attention'. In other words, `the more contacts you have and make, the more valuable you become, because people think you are popular and hence want to connect with you' \cite{Van_Dijck2013-xf}. 

How exactly scientific popularity influences regulation still needs to be explored in greater detail, particularly as a novel example of potential regulatory capture. What we know so far is that social media influencers can be highly effective in relying on authenticity and para-social relations for persuasion purposes \cite{Vannini2008-ii, Hudders2021-kv}.  In this context, popularity determines power relationships within social media networks that may capture regulatory processes in two ways. First, by exercising persuasion over policy-makers as audiences through visibility and popularity. Simply put, not all research that is available in a given field is presented on social media. Under the premise of basing policy on scientific evidence, politicians may rely on research that gains visibility due to amplification by scientific influencers, particularly when social media popularity is doubled by the brand power of prestigious academic institutions. Second, by amplifying polarized debates that may trigger policy options which are not sufficiently informed through transparent and collective processes of evidence gathering. If multiple AI research groups are vocal on social media about the future of AI research, this fuels a race towards AI regulation. This can take away from the thoroughness that is necessary in collecting evidence for such a complex field.

\section{Regulation and NLP (RegNLP): A New Field}

The danger of regulatory capture, taken together with the lack of systematization in the identification and measurement of risk and uncertainty around LLMs, calls for a cohesive scientific agenda and strategy. 
In 2023, the world counts 8 billion humans\footnote{\url{https://ourworldindata.org/world-population-growth.}For a brief comparison, at the time of the Dartmouth AI workshop in 1956, the world population was at a mere 2.5 billion, Statista, 2023.}, and further digitalization and automation are not only unstoppable, but also absolutely necessary. Technological innovation is currently vital in the governance of our society. That does not mean its pursuit ought to be free from regulatory frameworks mandating rules on how to deal with the risks it poses. NLP research has already drawn attention to some of the potential risks of LLMs. To consolidate this effort, it is necessary to consider in what direction NLP research can further develop and what contributions it can make to regulation. A concrete proposal we advocate for is the creation of a new field of scientific inquiry which we call Regulation and NLP (RegNLP). RegNLP has three essential features which we discuss below. 

\subsection{Multidisciplinarity}
First, RegNLP needs to be a multidisciplinary field that spans across any scientific areas of study which are relevant for the intersection of regulation and AI. In the past years, multidisciplinary communities have been increasingly popular. An example is the ACM Fairness, Accountability and Transparency Conference (FAccT\footnote{\url{https://facctconference.org}}), which often features NLP research. Such research communities form around the disciplines that are most reflected by their research questions. For RegNLP, the constitutive disciplines ought to include NLP and regulation studies but also law, economics, political science, etc.. NLP research approaches cannot replace expertise from other fields. At the same time, expertise entails more than an interest in an adjacent field, but rather a deep understanding of the contributions and limitations such a field can entail when interacting with NLP. One strategy to encourage this cross-pollination is for NLP researchers interested in regulation to co-author papers with regulation experts and other relevant scholars. In doing so, RegNLP can also contribute to the research gaps in other fields, such as public administration, where literature on AI still needs further development. For instance, in 2019, only 12 scientific articles were published on AI and public policy and administration, mostly focused on the use of AI \textit{in} public administration \cite{Valle-Cruz2020-ua}. Similarly, a quick search in `Regulation \& Governance', a leading journal in regulation studies, yields a total of 18 results, out of which only one discusses AI risks \cite{Laux2023-iq}.          

\subsection{Harmonized Methodologies}
RegNLP needs harmonized methodologies. One of the biggest problems with the consolidation of multidisciplinary research agendas and communities relates to the lack of alignment between the different methods and goals pursued by different researchers. This issue trickles down into all relevant activities which normally help consolidate multidisciplinary groups, and is most specifically visible in the process of peer review. If reviewers are not familiar with methodologies from other fields, they will be unable to adequately assess the quality of research \cite{Laufer}. This can lead to the publication of research which may be interesting across disciplines, but which may not meet the methodological rigor of the scientific discipline to which a given method pertains.  

RegNLP can help establish shared standards for scientific quality around shared methodologies and science practices. This can mean embracing a diverse methodological scope to reflect the tools that are needed in the inquiry of different types of research questions. It can also mean perfecting existing methods and deploying them on novel sources of data, as NLP research methods are a natural starting point for the systemic retrieval of complex information and overviews from existing scientific research, such as meta-studies \cite{Heijden2021-ju}. 

\subsection{Science Participation in Regulation}
RegNLP can help research on regulation and NLP interface with regulatory processes. At a time of increased complexity, it is important for scientists to clarify the state of art of fast paced technological change. Using harmonized methodologies in the context of a multidisciplinary research agenda can bring much needed coordination to the interaction between NLP development and regulation. In the absence of such coordination, as we have discussed in Section~\ref{sec:sci_expert_soc_media}, there is a potential danger that policy-makers are only exposed to popular scientific opinions instead of consolidated science communication. What is more, embedding RegNLP into a risk and regulation context can offer further inspiration for the role of academia, as a repository of public trust. A lot of regulatory agencies and standardization bodies govern the implementation of regulation. In addition, new forms of interactions with civil society are being set up by EU regulation, such as the Digital Services Act's `trusted flaggers', namely organizations that can flag illegal content on online platforms. Similarly, there can be new roles to play for RegNLP agendas and communities.  




\section{Conclusion}

LLMs reflect a momentous development in NLP research. As they unleash a new era of automation, it is important to understand their risks and how these risks can be controlled. While eager to engage with regulatory matters, NLP research on LLM risks has so far been disjointed from other fields which are of direct interest, such as regulation studies. In particular, the field of risk and uncertainty has been conceptualizing and discussing scientific risks for decades. In this paper, we introduced these two areas of study and explained why it would be beneficial for NLP research to consider them in greater depth. In doing so, we also raised concerns relating to the fact that a lot of scientific debates on NLP risks are taking place on social media. This may lead to regulatory capture, or in other words the exercise of influence over law-makers, who are notoriously active on social media platforms. To tackle these issues, we propose a new multidisciplinary area of scientific inquiry at the intersection of regulation and NLP (RegNLP), aimed at the development of a systematic approach for the identification and measurement of risks arising out of LLMs and NLP technology more broadly.  

\section*{Limitations}

Our paper reflects on the future of NLP in a landscape where interest in regulation is increasing exponentially, within and outside the field. Given the nature of this paper, we will refer to limitations dealing with the feasibility of our proposed research agenda. The most important limitation reflects the discussion around inter- and multidisciplinarity. This is by no means a new theme in science, and its implementation has cultural, managerial and economic implications that we do not discuss in the paper, but which are important to acknowledge. Similarly, another limitation is reflected by the modest amount of knowledge we have relating to the impact of social media influencers (such as science influencers) on regulation and public policy. In this paper, we raise certain issues around science communication as the starting point of a broader discussion around power and influence in law-making as amplified by social media.

\bibliography{anthology,custom}

\begin{thebibliography}{79}
\expandafter\ifx\csname natexlab\endcsname\relax\def\natexlab#1{#1}\fi

\bibitem[{Abdalla and Abdalla(2021)}]{abdalla2021hoodie}
Mohamed Abdalla and Moustafa Abdalla. 2021.
\newblock \href {https://doi.org/10.1145/3461702.3462563} {The grey hoodie
  project: Big tobacco, big tech, and the threat on academic integrity}.
\newblock In \emph{Proceedings of the 2021 AAAI/ACM Conference on AI, Ethics,
  and Society}, AIES '21, page 287–297, New York, NY, USA. Association for
  Computing Machinery.

\bibitem[{Abdalla et~al.(2023)Abdalla, Wahle, Ruas, Névéol, Ducel, Mohammad,
  and Fort}]{abdalla2023elephant}
Mohamed Abdalla, Jan~Philip Wahle, Terry Ruas, Aurélie Névéol, Fanny Ducel,
  Saif~M. Mohammad, and Karën Fort. 2023.
\newblock \href {http://arxiv.org/abs/2305.02797} {The elephant in the room:
  Analyzing the presence of big tech in natural language processing research}.

\bibitem[{Anil et~al.(2023)Anil, Dai, Firat, Johnson, Lepikhin, Passos,
  Shakeri, Taropa, Bailey, Chen, Chu, Clark, Shafey, Huang, Meier-Hellstern,
  Mishra, Moreira, Omernick, Robinson, Ruder, Tay, Xiao, Xu, Zhang, Abrego,
  Ahn, Austin, Barham, Botha, Bradbury, Brahma, Brooks, Catasta, Cheng, Cherry,
  Choquette-Choo, Chowdhery, Crepy, Dave, Dehghani, Dev, Devlin, Díaz, Du,
  Dyer, Feinberg, Feng, Fienber, Freitag, Garcia, Gehrmann, Gonzalez, Gur-Ari,
  Hand, Hashemi, Hou, Howland, Hu, Hui, Hurwitz, Isard, Ittycheriah, Jagielski,
  Jia, Kenealy, Krikun, Kudugunta, Lan, Lee, Lee, Li, Li, Li, Li, Li, Lim, Lin,
  Liu, Liu, Maggioni, Mahendru, Maynez, Misra, Moussalem, Nado, Nham, Ni,
  Nystrom, Parrish, Pellat, Polacek, Polozov, Pope, Qiao, Reif, Richter, Riley,
  Ros, Roy, Saeta, Samuel, Shelby, Slone, Smilkov, So, Sohn, Tokumine, Valter,
  Vasudevan, Vodrahalli, Wang, Wang, Wang, Wang, Wieting, Wu, Xu, Xu, Xue, Yin,
  Yu, Zhang, Zheng, Zheng, Zhou, Zhou, Petrov, and Wu}]{anil2023palm}
Rohan Anil, Andrew~M. Dai, Orhan Firat, Melvin Johnson, Dmitry Lepikhin,
  Alexandre Passos, Siamak Shakeri, Emanuel Taropa, Paige Bailey, Zhifeng Chen,
  Eric Chu, Jonathan~H. Clark, Laurent~El Shafey, Yanping Huang, Kathy
  Meier-Hellstern, Gaurav Mishra, Erica Moreira, Mark Omernick, Kevin Robinson,
  Sebastian Ruder, Yi~Tay, Kefan Xiao, Yuanzhong Xu, Yujing Zhang,
  Gustavo~Hernandez Abrego, Junwhan Ahn, Jacob Austin, Paul Barham, Jan Botha,
  James Bradbury, Siddhartha Brahma, Kevin Brooks, Michele Catasta, Yong Cheng,
  Colin Cherry, Christopher~A. Choquette-Choo, Aakanksha Chowdhery, Clément
  Crepy, Shachi Dave, Mostafa Dehghani, Sunipa Dev, Jacob Devlin, Mark Díaz,
  Nan Du, Ethan Dyer, Vlad Feinberg, Fangxiaoyu Feng, Vlad Fienber, Markus
  Freitag, Xavier Garcia, Sebastian Gehrmann, Lucas Gonzalez, Guy Gur-Ari,
  Steven Hand, Hadi Hashemi, Le~Hou, Joshua Howland, Andrea Hu, Jeffrey Hui,
  Jeremy Hurwitz, Michael Isard, Abe Ittycheriah, Matthew Jagielski, Wenhao
  Jia, Kathleen Kenealy, Maxim Krikun, Sneha Kudugunta, Chang Lan, Katherine
  Lee, Benjamin Lee, Eric Li, Music Li, Wei Li, YaGuang Li, Jian Li, Hyeontaek
  Lim, Hanzhao Lin, Zhongtao Liu, Frederick Liu, Marcello Maggioni, Aroma
  Mahendru, Joshua Maynez, Vedant Misra, Maysam Moussalem, Zachary Nado, John
  Nham, Eric Ni, Andrew Nystrom, Alicia Parrish, Marie Pellat, Martin Polacek,
  Alex Polozov, Reiner Pope, Siyuan Qiao, Emily Reif, Bryan Richter, Parker
  Riley, Alex~Castro Ros, Aurko Roy, Brennan Saeta, Rajkumar Samuel, Renee
  Shelby, Ambrose Slone, Daniel Smilkov, David~R. So, Daniel Sohn, Simon
  Tokumine, Dasha Valter, Vijay Vasudevan, Kiran Vodrahalli, Xuezhi Wang,
  Pidong Wang, Zirui Wang, Tao Wang, John Wieting, Yuhuai Wu, Kelvin Xu, Yunhan
  Xu, Linting Xue, Pengcheng Yin, Jiahui Yu, Qiao Zhang, Steven Zheng,
  Ce~Zheng, Weikang Zhou, Denny Zhou, Slav Petrov, and Yonghui Wu. 2023.
\newblock \href {http://arxiv.org/abs/2305.10403} {Palm 2 technical report}.

\bibitem[{Aumann and Serrano(2008)}]{Aumann2008-lk}
Robert~J Aumann and Roberto Serrano. 2008.
\newblock An economic index of riskiness.
\newblock \emph{J. Polit. Econ.}, 116(5):810--836.

\bibitem[{Barocas et~al.(2019)Barocas, Hardt, and
  Narayanan}]{barocas-hardt-narayanan}
Solon Barocas, Moritz Hardt, and Arvind Narayanan. 2019.
\newblock \emph{Fairness and Machine Learning: Limitations and Opportunities}.
\newblock fairmlbook.org.
\newblock \url{http://www.fairmlbook.org}.

\bibitem[{Bender et~al.(2021)Bender, Gebru, McMillan-Major, and
  Shmitchell}]{Bender2021}
Emily~M. Bender, Timnit Gebru, Angelina McMillan-Major, and Shmargaret
  Shmitchell. 2021.
\newblock \href {https://doi.org/10.1145/3442188.3445922} {On the dangers of
  stochastic parrots: Can language models be too big?}
\newblock In \emph{Proceedings of the 2021 ACM Conference on Fairness,
  Accountability, and Transparency}, FAccT '21, page 610–623, New York, NY,
  USA. Association for Computing Machinery.

\bibitem[{Birhane et~al.(2021)Birhane, Prabhu, and
  Kahembwe}]{birhane2021multimodal}
Abeba Birhane, Vinay~Uday Prabhu, and Emmanuel Kahembwe. 2021.
\newblock \href {http://arxiv.org/abs/2110.01963} {Multimodal datasets:
  misogyny, pornography, and malignant stereotypes}.

\bibitem[{Black(2008)}]{Black2008-uk}
Julia Black. 2008.
\newblock Constructing and contesting legitimacy and accountability in
  polycentric regulatory regimes.
\newblock \emph{Regul. Gov.}, 2(2):137--164.

\bibitem[{Blodgett et~al.(2020)Blodgett, Barocas, Daum{\'e}~III, and
  Wallach}]{blodgett-etal-2020-language}
Su~Lin Blodgett, Solon Barocas, Hal Daum{\'e}~III, and Hanna Wallach. 2020.
\newblock \href {https://doi.org/10.18653/v1/2020.acl-main.485} {Language
  (technology) is power: A critical survey of {``}bias{''} in {NLP}}.
\newblock In \emph{Proceedings of the 58th Annual Meeting of the Association
  for Computational Linguistics}, pages 5454--5476, Online. Association for
  Computational Linguistics.

\bibitem[{Bommasani et~al.(2022)Bommasani, Hudson, Adeli, Altman, Arora, von
  Arx, Bernstein, Bohg, Bosselut, Brunskill, Brynjolfsson, Buch, Card,
  Castellon, Chatterji, Chen, Creel, Davis, Demszky, Donahue, Doumbouya,
  Durmus, Ermon, Etchemendy, Ethayarajh, Fei-Fei, Finn, Gale, Gillespie, Goel,
  Goodman, Grossman, Guha, Hashimoto, Henderson, Hewitt, Ho, Hong, Hsu, Huang,
  Icard, Jain, Jurafsky, Kalluri, Karamcheti, Keeling, Khani, Khattab, Koh,
  Krass, Krishna, Kuditipudi, Kumar, Ladhak, Lee, Lee, Leskovec, Levent, Li,
  Li, Ma, Malik, Manning, Mirchandani, Mitchell, Munyikwa, Nair, Narayan,
  Narayanan, Newman, Nie, Niebles, Nilforoshan, Nyarko, Ogut, Orr,
  Papadimitriou, Park, Piech, Portelance, Potts, Raghunathan, Reich, Ren, Rong,
  Roohani, Ruiz, Ryan, Ré, Sadigh, Sagawa, Santhanam, Shih, Srinivasan,
  Tamkin, Taori, Thomas, Tramèr, Wang, Wang, Wu, Wu, Wu, Xie, Yasunaga, You,
  Zaharia, Zhang, Zhang, Zhang, Zhang, Zheng, Zhou, and
  Liang}]{bommasani2022opportunities}
Rishi Bommasani, Drew~A. Hudson, Ehsan Adeli, Russ Altman, Simran Arora, Sydney
  von Arx, Michael~S. Bernstein, Jeannette Bohg, Antoine Bosselut, Emma
  Brunskill, Erik Brynjolfsson, Shyamal Buch, Dallas Card, Rodrigo Castellon,
  Niladri Chatterji, Annie Chen, Kathleen Creel, Jared~Quincy Davis, Dora
  Demszky, Chris Donahue, Moussa Doumbouya, Esin Durmus, Stefano Ermon, John
  Etchemendy, Kawin Ethayarajh, Li~Fei-Fei, Chelsea Finn, Trevor Gale, Lauren
  Gillespie, Karan Goel, Noah Goodman, Shelby Grossman, Neel Guha, Tatsunori
  Hashimoto, Peter Henderson, John Hewitt, Daniel~E. Ho, Jenny Hong, Kyle Hsu,
  Jing Huang, Thomas Icard, Saahil Jain, Dan Jurafsky, Pratyusha Kalluri,
  Siddharth Karamcheti, Geoff Keeling, Fereshte Khani, Omar Khattab, Pang~Wei
  Koh, Mark Krass, Ranjay Krishna, Rohith Kuditipudi, Ananya Kumar, Faisal
  Ladhak, Mina Lee, Tony Lee, Jure Leskovec, Isabelle Levent, Xiang~Lisa Li,
  Xuechen Li, Tengyu Ma, Ali Malik, Christopher~D. Manning, Suvir Mirchandani,
  Eric Mitchell, Zanele Munyikwa, Suraj Nair, Avanika Narayan, Deepak
  Narayanan, Ben Newman, Allen Nie, Juan~Carlos Niebles, Hamed Nilforoshan,
  Julian Nyarko, Giray Ogut, Laurel Orr, Isabel Papadimitriou, Joon~Sung Park,
  Chris Piech, Eva Portelance, Christopher Potts, Aditi Raghunathan, Rob Reich,
  Hongyu Ren, Frieda Rong, Yusuf Roohani, Camilo Ruiz, Jack Ryan, Christopher
  Ré, Dorsa Sadigh, Shiori Sagawa, Keshav Santhanam, Andy Shih, Krishnan
  Srinivasan, Alex Tamkin, Rohan Taori, Armin~W. Thomas, Florian Tramèr,
  Rose~E. Wang, William Wang, Bohan Wu, Jiajun Wu, Yuhuai Wu, Sang~Michael Xie,
  Michihiro Yasunaga, Jiaxuan You, Matei Zaharia, Michael Zhang, Tianyi Zhang,
  Xikun Zhang, Yuhui Zhang, Lucia Zheng, Kaitlyn Zhou, and Percy Liang. 2022.
\newblock \href {http://arxiv.org/abs/2108.07258} {On the opportunities and
  risks of foundation models}.

\bibitem[{Borenstein et~al.(2021)Borenstein, Grodzinsky, Howard, Miller, and
  Wolf}]{Borenstein2021-ws}
Jason Borenstein, Frances~S Grodzinsky, Ayanna Howard, Keith~W Miller, and
  Marty~J Wolf. 2021.
\newblock {AI} ethics: A long history and a recent burst of attention.
\newblock \emph{Computer (Long Beach Calif.)}, 54(1):96--102.

\bibitem[{Brownsword et~al.(2017)Brownsword, Scotford, and
  Yeung}]{Brownsword2017-gq}
Roger Brownsword, Eloise Scotford, and Karen Yeung, editors. 2017.
\newblock \emph{The oxford handbook of law, regulation and technology}.
\newblock Oxford Handbooks. Oxford University Press, London, England.

\bibitem[{Carpenter and Moss(2013)}]{Carpenter2013-bv}
Daniel Carpenter and David~A Moss, editors. 2013.
\newblock \emph{Preventing regulatory capture}.
\newblock Cambridge University Press, Cambridge, England.

\bibitem[{Christiano et~al.(2017)Christiano, Leike, Brown, Martic, Legg, and
  Amodei}]{christiano-etal-2017-drl}
Paul Christiano, Jan Leike, Tom~B. Brown, Miljan Martic, Shane Legg, and Dario
  Amodei. 2017.
\newblock \href {https://doi.org/10.48550/ARXIV.1706.03741} {Deep reinforcement
  learning from human preferences}.

\bibitem[{Chung et~al.(2022)Chung, Hou, Longpre, Zoph, Tay, Fedus, Li, Wang,
  Dehghani, Brahma, Webson, Gu, Dai, Suzgun, Chen, Chowdhery, Castro-Ros,
  Pellat, Robinson, Valter, Narang, Mishra, Yu, Zhao, Huang, Dai, Yu, Petrov,
  Chi, Dean, Devlin, Roberts, Zhou, Le, and Wei}]{chung2022scaling}
Hyung~Won Chung, Le~Hou, Shayne Longpre, Barret Zoph, Yi~Tay, William Fedus,
  Yunxuan Li, Xuezhi Wang, Mostafa Dehghani, Siddhartha Brahma, Albert Webson,
  Shixiang~Shane Gu, Zhuyun Dai, Mirac Suzgun, Xinyun Chen, Aakanksha
  Chowdhery, Alex Castro-Ros, Marie Pellat, Kevin Robinson, Dasha Valter,
  Sharan Narang, Gaurav Mishra, Adams Yu, Vincent Zhao, Yanping Huang, Andrew
  Dai, Hongkun Yu, Slav Petrov, Ed~H. Chi, Jeff Dean, Jacob Devlin, Adam
  Roberts, Denny Zhou, Quoc~V. Le, and Jason Wei. 2022.
\newblock \href {http://arxiv.org/abs/2210.11416} {Scaling
  instruction-finetuned language models}.

\bibitem[{Commission(2023{\natexlab{a}})}]{Commission2022}
European Commission. 2023{\natexlab{a}}.
\newblock Artificial intelligence – ethical and legal requirements.
\newblock
  \url{https://ec.europa.eu/info/law/better-regulation/have-your-say/initiatives/12527-Artificial-intelligence-ethical-and-legal-requirements/feedback_en?p_id=24212003}.
\newblock [Accessed 16-Jun-2023].

\bibitem[{Commission(2023{\natexlab{b}})}]{Commission2023}
European Commission. 2023{\natexlab{b}}.
\newblock Better regulation: why and how.
\newblock
  \url{https://commission.europa.eu/law/law-making-process/planning-and-proposing-law/better-regulation_en#:~:text=The%20Better%20Regulation%20agenda%20ensures,those%20that%20may%20be%20affected}.
\newblock [Accessed 16-Jun-2023].

\bibitem[{Dal~Bo(2006)}]{Dal_Bo2006-hn}
E~Dal~Bo. 2006.
\newblock Regulatory capture: A review.
\newblock \emph{Oxf. Rev. Econ. Pol.}, 22(2):203--225.

\bibitem[{Derczynski et~al.(2023)Derczynski, Kirk, Balachandran, Kumar,
  Tsvetkov, Leiser, and Mohammad}]{derczynski2023assessing}
Leon Derczynski, Hannah~Rose Kirk, Vidhisha Balachandran, Sachin Kumar, Yulia
  Tsvetkov, MR~Leiser, and Saif Mohammad. 2023.
\newblock Assessing language model deployment with risk cards.
\newblock \emph{arXiv preprint arXiv:2303.18190}.

\bibitem[{Dobbe et~al.(2021)Dobbe, Krendl~Gilbert, and Mintz}]{Dobbe2021-wa}
Roel Dobbe, Thomas Krendl~Gilbert, and Yonatan Mintz. 2021.
\newblock Hard choices in artificial intelligence.
\newblock \emph{Artif. Intell.}, 300(103555):103555.

\bibitem[{Field et~al.(2023)Field, Coston, Gandhi, Chouldechova,
  Putnam-Hornstein, Steier, and Tsvetkov}]{Field2023-de}
Anjalie Field, Amanda Coston, Nupoor Gandhi, Alexandra Chouldechova, Emily
  Putnam-Hornstein, David Steier, and Yulia Tsvetkov. 2023.
\newblock Examining risks of racial biases in {NLP} tools for child protective
  services.
\newblock In \emph{2023 {ACM} Conference on Fairness, Accountability, and
  Transparency}, New York, NY, USA. ACM.

\bibitem[{Floridi(2023)}]{Floridi2023-vl}
Luciano Floridi. 2023.
\newblock \emph{The ethics of artificial intelligence the ethics of artificial
  intelligence}.
\newblock Oxford University Press, London, England.

\bibitem[{Gabriel(2020)}]{gabriel-2020-alignment}
Iason Gabriel. 2020.
\newblock \href {https://doi.org/10.1007/s11023-020-09539-2} {Artificial
  intelligence, values, and alignment}.
\newblock \emph{Minds Mach.}, 30(3):411–437.

\bibitem[{Garber(2012)}]{Garber}
Megan Garber. 2012.
\newblock {T}he {I}{V}{F} {P}anic: '{A}ll {H}ell {W}ill {B}reak {L}oose,
  {P}olitically and {M}orally, {A}ll {O}ver the {W}orld' --- theatlantic.com.
\newblock
  \url{https://www.theatlantic.com/technology/archive/2012/06/the-ivf-panic-all-hell-will-break-loose-politically-and-morally-all-over-the-world/258954/}.
\newblock [Accessed 16-Jun-2023].

\bibitem[{Genus and Stirling(2018)}]{Genus2018-uq}
Audley Genus and Andy Stirling. 2018.
\newblock Collingridge and the dilemma of control: Towards responsible and
  accountable innovation.
\newblock \emph{Res. Policy}, 47(1):61--69.

\bibitem[{Goanta and Ranchordás(2020)}]{Goanta_Ranchordás_2020}
Catalina Goanta and Sofia Ranchordás. 2020.
\newblock \emph{The regulation of Social Media Influencers}.
\newblock Edward Elgar Publishing.

\bibitem[{Gough(1988)}]{Gough}
Janet Gough. 1988.
\newblock \href
  {https://researcharchive.lincoln.ac.nz/bitstream/handle/10182/1376/crm_ip_10.pdf\%3Bsequence\%3D1}
  {Risk and uncertainty}.
\newblock [Accessed 16-Jun-2023].

\bibitem[{Heijden(2021)}]{Heijden2021-ju}
Jeroen Heijden. 2021.
\newblock Why meta‐research matters to regulation and governance scholarship:
  An illustrative evidence synthesis of responsive regulation research.
\newblock \emph{Regul. Gov.}, 15(S1).

\bibitem[{Hemphill(2020)}]{Hemphill2020-gt}
Thomas~A Hemphill. 2020.
\newblock ``the innovation governance dilemma: Alternatives to the
  precautionary principle''.
\newblock \emph{Technol. Soc.}, 63(101381):101381.

\bibitem[{Hendrycks et~al.(2023)Hendrycks, Mazeika, and
  Woodside}]{hendrycks2023overview}
Dan Hendrycks, Mantas Mazeika, and Thomas Woodside. 2023.
\newblock \href {http://arxiv.org/abs/2306.12001} {An overview of catastrophic
  ai risks}.

\bibitem[{Heyvaert(2011)}]{Heyvaert2011-bf}
Veerle Heyvaert. 2011.
\newblock Governing climate change: Towards a new paradigm for risk regulation.
\newblock \emph{Mod. Law Rev.}, 74(6):817--844.

\bibitem[{Hovy and Prabhumoye(2021)}]{Hovy2021-wt}
Dirk Hovy and Shrimai Prabhumoye. 2021.
\newblock Five sources of bias in natural language processing.
\newblock \emph{Lang. Linguist. Compass}, 15(8):e12432.

\bibitem[{Howarth(2013)}]{Howarth2013-rj}
Anita Howarth. 2013.
\newblock A `superstorm': when moral panic and new risk discourses converge in
  the media.
\newblock \emph{Health Risk Soc.}, 15(8):681--698.

\bibitem[{Hudders et~al.(2021)Hudders, De~Jans, and
  De~Veirman}]{Hudders2021-kv}
Liselot Hudders, Steffi De~Jans, and Marijke De~Veirman. 2021.
\newblock The commercialization of social media stars: a literature review and
  conceptual framework on the strategic use of social media influencers.
\newblock \emph{Int. J. Advert.}, 40(3):327--375.

\bibitem[{Kidd and Birhane(2023)}]{Kidd2023-iw}
Celeste Kidd and Abeba Birhane. 2023.
\newblock How {AI} can distort human beliefs.
\newblock \emph{Science}, 380(6651):1222--1223.

\bibitem[{Kim and Song(2016)}]{Kim2016-ie}
Jihyun Kim and Hayeon Song. 2016.
\newblock Celebrity's self-disclosure on twitter and parasocial relationships:
  A mediating role of social presence.
\newblock \emph{Comput. Human Behav.}, 62:570--577.

\bibitem[{Krebs(2011)}]{Krebs}
John~R. Krebs. 2011.
\newblock \href {https://www.jstor.org/stable/23057221} {Handling uncertainty
  in science}.
\newblock \emph{Philosophical Transactions: Mathematical, Physical and
  Engineering Sciences}, (1956):4842--4852.

\bibitem[{Laufer et~al.(2022)Laufer, Jain, Cooper, Kleinberg, and
  Heidari}]{Laufer}
Benjamin Laufer, Sameer Jain, A.~Feder Cooper, Jon Kleinberg, and Hoda Heidari.
  2022.
\newblock \href {https://doi.org/10.1145/3531146.3533107} {Four years of facct:
  A reflexive, mixed-methods analysis of research contributions, shortcomings,
  and future prospects}.
\newblock In \emph{Proceedings of the 2022 ACM Conference on Fairness,
  Accountability, and Transparency}, FAccT '22, page 401–426, New York, NY,
  USA. Association for Computing Machinery.

\bibitem[{Laux et~al.(2023)Laux, Wachter, and Mittelstadt}]{Laux2023-iq}
Johann Laux, Sandra Wachter, and Brent Mittelstadt. 2023.
\newblock Trustworthy artificial intelligence and the european union {AI} act:
  On the conflation of trustworthiness and acceptability of risk.
\newblock \emph{Regul. Gov.}

\bibitem[{Leike et~al.(2018)Leike, Krueger, Everitt, Martic, Maini, and
  Legg}]{leike-etal-2018-alignnment}
Jan Leike, David Krueger, Tom Everitt, Miljan Martic, Vishal Maini, and Shane
  Legg. 2018.
\newblock \href {http://arxiv.org/abs/1811.07871} {Scalable agent alignment via
  reward modeling: a research direction}.
\newblock \emph{CoRR}, abs/1811.07871.

\bibitem[{Levine and Forrence(1990)}]{levine}
Michael~E. Levine and Jennifer~L. Forrence. 1990.
\newblock \href {http://www.jstor.org/stable/764987} {Regulatory capture,
  public interest, and the public agenda: Toward a synthesis}.
\newblock \emph{Journal of Law, Economics, \& Organization}, 6:167--198.

\bibitem[{Marchant et~al.(2013)Marchant, Allenby, and Herkert}]{marchant}
Gary~E. Marchant, Braden~R. Allenby, and Joseph~R. Herkert. 2013.
\newblock \emph{The Growing Gap Between Emerging Technologies and Legal-Ethical
  Oversight: The Pacing Problem}.
\newblock Springer Publishing Company, Incorporated.

\bibitem[{Mireshghallah et~al.(2022)Mireshghallah, Goyal, Uniyal,
  Berg-Kirkpatrick, and Shokri}]{mireshghallah-etal-2022-quantifying}
Fatemehsadat Mireshghallah, Kartik Goyal, Archit Uniyal, Taylor
  Berg-Kirkpatrick, and Reza Shokri. 2022.
\newblock \href {https://aclanthology.org/2022.emnlp-main.570} {Quantifying
  privacy risks of masked language models using membership inference attacks}.
\newblock In \emph{Proceedings of the 2022 Conference on Empirical Methods in
  Natural Language Processing}, pages 8332--8347, Abu Dhabi, United Arab
  Emirates. Association for Computational Linguistics.

\bibitem[{Morgan and Yeung(2007)}]{Morgan2007-he}
Bronwen Morgan and Karen Yeung. 2007.
\newblock \emph{An introduction to law and regulation: Text and materials}.
\newblock Cambridge University Press, Cambridge, England.

\bibitem[{Ogus(2009)}]{Ogus}
Anthony Ogus. 2009.
\newblock Regulation revisited.
\newblock \emph{Public Law}, (2):332--346.

\bibitem[{OpenAI(2023)}]{openai2023gpt4}
OpenAI. 2023.
\newblock \href {http://arxiv.org/abs/2303.08774} {Gpt-4 technical report}.

\bibitem[{Ouyang et~al.(2022)Ouyang, Wu, Jiang, Almeida, Wainwright, Mishkin,
  Zhang, Agarwal, Slama, Ray, Schulman, Hilton, Kelton, Miller, Simens, Askell,
  Welinder, Christiano, Leike, and Lowe}]{instructgpt}
Long Ouyang, Jeff Wu, Xu~Jiang, Diogo Almeida, Carroll~L. Wainwright, Pamela
  Mishkin, Chong Zhang, Sandhini Agarwal, Katarina Slama, Alex Ray, John
  Schulman, Jacob Hilton, Fraser Kelton, Luke Miller, Maddie Simens, Amanda
  Askell, Peter Welinder, Paul Christiano, Jan Leike, and Ryan Lowe. 2022.
\newblock \href {https://doi.org/10.48550/ARXIV.2203.02155} {Training language
  models to follow instructions with human feedback}.

\bibitem[{Page et~al.(2021)Page, McKenzie, Bossuyt, Boutron, Hoffmann, Mulrow,
  Shamseer, Tetzlaff, Akl, Brennan, Chou, Glanville, Grimshaw,
  Hr{\'o}bjartsson, Lalu, Li, Loder, Mayo-Wilson, McDonald, McGuinness,
  Stewart, Thomas, Tricco, Welch, Whiting, and Moher}]{Page2021-kx}
Matthew~J Page, Joanne~E McKenzie, Patrick~M Bossuyt, Isabelle Boutron, Tammy~C
  Hoffmann, Cynthia~D Mulrow, Larissa Shamseer, Jennifer~M Tetzlaff, Elie~A
  Akl, Sue~E Brennan, Roger Chou, Julie Glanville, Jeremy~M Grimshaw,
  Asbj{\o}rn Hr{\'o}bjartsson, Manoj~M Lalu, Tianjing Li, Elizabeth~W Loder,
  Evan Mayo-Wilson, Steve McDonald, Luke~A McGuinness, Lesley~A Stewart, James
  Thomas, Andrea~C Tricco, Vivian~A Welch, Penny Whiting, and David Moher.
  2021.
\newblock The {PRISMA} 2020 statement: an updated guideline for reporting
  systematic reviews.
\newblock \emph{Syst. Rev.}, 10(1):89.

\bibitem[{Park and Shapira(2017)}]{Park2017-og}
K~Francis Park and Zur Shapira. 2017.
\newblock Risk and uncertainty.
\newblock In \emph{The Palgrave Encyclopedia of Strategic Management}, pages
  1--7. Palgrave Macmillan UK, London.

\bibitem[{Paschke et~al.(2019)Paschke, Pfisterer, Hirschi, Last, Pauli, Studer,
  Schubert, Herrend{\"o}rfer, and Mc~Nally}]{Paschke2019-nr}
Melanie Paschke, Andrea Pfisterer, Christian Hirschi, Luisa Last, Daniela
  Pauli, Bruno Studer, Jasmin Schubert, Robert Herrend{\"o}rfer, and
  Kaitlin~Elyse Mc~Nally. 2019.
\newblock Evidence-based policymaking.

\bibitem[{Porter(2020)}]{Porter2020-zx}
Theodore~M Porter. 2020.
\newblock Objectivity and the politics of disciplines.
\newblock In \emph{Trust in Numbers}, pages 193--216. Princeton University
  Press.

\bibitem[{Radford et~al.(2018)Radford, Narasimhan, Salimans, and
  Sutskever}]{radford2018improving}
Alec Radford, Karthik Narasimhan, Tim Salimans, and Ilya Sutskever. 2018.
\newblock Improving language understanding by generative pre-training.

\bibitem[{Riedl et~al.(2023)Riedl, Lukito, and Woolley}]{Riedl2023-zc}
Martin~J Riedl, Josephine Lukito, and Samuel~C Woolley. 2023.
\newblock Political influencers on social media: An introduction.
\newblock \emph{Soc. Media Soc.}, 9(2):205630512311779.

\bibitem[{Rillig et~al.(2023)Rillig, {\AA}gerstrand, Bi, Gould, and
  Sauerland}]{Rillig2023-kp}
Matthias~C Rillig, Marlene {\AA}gerstrand, Mohan Bi, Kenneth~A Gould, and Uli
  Sauerland. 2023.
\newblock Risks and benefits of large language models for the environment.
\newblock \emph{Environ. Sci. Technol.}, 57(9):3464--3466.

\bibitem[{Rolnick et~al.(2022)Rolnick, Donti, Kaack, Kochanski, Lacoste,
  Sankaran, Ross, Milojevic-Dupont, Jaques, Waldman-Brown, Luccioni, Maharaj,
  Sherwin, Mukkavilli, Kording, Gomes, Ng, Hassabis, Platt, Creutzig, Chayes,
  and Bengio}]{rolnick}
David Rolnick, Priya~L. Donti, Lynn~H. Kaack, Kelly Kochanski, Alexandre
  Lacoste, Kris Sankaran, Andrew~Slavin Ross, Nikola Milojevic-Dupont, Natasha
  Jaques, Anna Waldman-Brown, Alexandra~Sasha Luccioni, Tegan Maharaj, Evan~D.
  Sherwin, S.~Karthik Mukkavilli, Konrad~P. Kording, Carla~P. Gomes, Andrew~Y.
  Ng, Demis Hassabis, John~C. Platt, Felix Creutzig, Jennifer Chayes, and
  Yoshua Bengio. 2022.
\newblock \href {https://doi.org/10.1145/3485128} {Tackling climate change with
  machine learning}.
\newblock \emph{ACM Comput. Surv.}, 55(2).

\bibitem[{Roose(2023)}]{nytimesAIPoses}
Kevin Roose. 2023.
\newblock {A}.{I}. {P}oses ‘{R}isk of {E}xtinction,’ {I}ndustry {L}eaders
  {W}arn --- nytimes.com.
\newblock
  \url{https://www.nytimes.com/2023/05/30/technology/ai-threat-warning.html}.
\newblock [Accessed 04-Jun-2023].

\bibitem[{Saltelli et~al.(2022)Saltelli, Dankel, Di~Fiore, Holland, and
  Pigeon}]{Saltelli2022}
Andrea Saltelli, Dorothy~J Dankel, Monica Di~Fiore, Nina Holland, and Martin
  Pigeon. 2022.
\newblock Science, the endless frontier of regulatory capture.
\newblock \emph{Futures}, 135(102860):102860.

\bibitem[{Schwartz et~al.(2020)Schwartz, Dodge, Smith, and
  Etzioni}]{Schwartz2020-df}
Roy Schwartz, Jesse Dodge, Noah~A Smith, and Oren Etzioni. 2020.
\newblock Green {AI}.
\newblock \emph{Commun. ACM}, 63(12):54--63.

\bibitem[{Selznick(1985)}]{selznick1985focusing}
Philip Selznick. 1985.
\newblock Focusing organizational research on regulation.
\newblock \emph{Regulatory policy and the social sciences}, 1(1):363--367.

\bibitem[{Simonelli and Iacob(2021)}]{simonelli_iacob_2021}
Felice Simonelli and Nadina Iacob. 2021.
\newblock \href {https://doi.org/10.1017/err.2021.40} {Can we better the
  european union better regulation agenda?}
\newblock \emph{European Journal of Risk Regulation}, 12(4):849–860.

\bibitem[{Stiennon et~al.(2020)Stiennon, Ouyang, Wu, Ziegler, Lowe, Voss,
  Radford, Amodei, and Christiano}]{Stiennon2020}
Nisan Stiennon, Long Ouyang, Jeffrey Wu, Daniel Ziegler, Ryan Lowe, Chelsea
  Voss, Alec Radford, Dario Amodei, and Paul~F Christiano. 2020.
\newblock \href
  {https://proceedings.neurips.cc/paper/2020/file/1f89885d556929e98d3ef9b86448f951-Paper.pdf}
  {Learning to summarize with human feedback}.
\newblock In \emph{Advances in Neural Information Processing Systems},
  volume~33, pages 3008--3021. Curran Associates, Inc.

\bibitem[{Strubell et~al.(2019)Strubell, Ganesh, and
  McCallum}]{strubell-etal-2019-energy}
Emma Strubell, Ananya Ganesh, and Andrew McCallum. 2019.
\newblock \href {https://doi.org/10.18653/v1/P19-1355} {Energy and policy
  considerations for deep learning in {NLP}}.
\newblock In \emph{Proceedings of the 57th Annual Meeting of the Association
  for Computational Linguistics}, pages 3645--3650, Florence, Italy.
  Association for Computational Linguistics.

\bibitem[{Talat et~al.(2022)Talat, N{\'e}v{\'e}ol, Biderman, Clinciu, Dey,
  Longpre, Luccioni, Masoud, Mitchell, Radev, Sharma, Subramonian, Tae, Tan,
  Tunuguntla, and Van Der~Wal}]{talat-etal-2022-reap}
Zeerak Talat, Aur{\'e}lie N{\'e}v{\'e}ol, Stella Biderman, Miruna Clinciu,
  Manan Dey, Shayne Longpre, Sasha Luccioni, Maraim Masoud, Margaret Mitchell,
  Dragomir Radev, Shanya Sharma, Arjun Subramonian, Jaesung Tae, Samson Tan,
  Deepak Tunuguntla, and Oskar Van Der~Wal. 2022.
\newblock \href {https://doi.org/10.18653/v1/2022.bigscience-1.3} {You reap
  what you sow: On the challenges of bias evaluation under multilingual
  settings}.
\newblock In \emph{Proceedings of BigScience Episode {\#}5 -- Workshop on
  Challenges {\&} Perspectives in Creating Large Language Models}, pages
  26--41, virtual+Dublin. Association for Computational Linguistics.

\bibitem[{Touvron et~al.(2023)Touvron, Lavril, Izacard, Martinet, Lachaux,
  Lacroix, Rozière, Goyal, Hambro, Azhar, Rodriguez, Joulin, Grave, and
  Lample}]{touvron2023llama}
Hugo Touvron, Thibaut Lavril, Gautier Izacard, Xavier Martinet, Marie-Anne
  Lachaux, Timothée Lacroix, Baptiste Rozière, Naman Goyal, Eric Hambro,
  Faisal Azhar, Aurelien Rodriguez, Armand Joulin, Edouard Grave, and Guillaume
  Lample. 2023.
\newblock \href {http://arxiv.org/abs/2302.13971} {Llama: Open and efficient
  foundation language models}.

\bibitem[{Tsarapatsanis and Aletras(2021)}]{tsarapatsanis-aletras-2021-ethical}
Dimitrios Tsarapatsanis and Nikolaos Aletras. 2021.
\newblock \href {https://doi.org/10.18653/v1/2021.findings-acl.314} {On the
  ethical limits of natural language processing on legal text}.
\newblock In \emph{Findings of the Association for Computational Linguistics:
  ACL-IJCNLP 2021}, pages 3590--3599, Online. Association for Computational
  Linguistics.

\bibitem[{Valle-Cruz et~al.(2020)Valle-Cruz, Criado, Sandoval-Almaz{\'a}n, and
  Ruvalcaba-Gomez}]{Valle-Cruz2020-ua}
David Valle-Cruz, J~Ignacio Criado, Rodrigo Sandoval-Almaz{\'a}n, and Edgar~A
  Ruvalcaba-Gomez. 2020.
\newblock Assessing the public policy-cycle framework in the age of artificial
  intelligence: From agenda-setting to policy evaluation.
\newblock \emph{Gov. Inf. Q.}, 37(4):101509.

\bibitem[{van~den Hoven(2014)}]{Van_den_Hoven2014-xn}
Jeroen van~den Hoven. 2014.
\newblock Responsible innovation: A new look at technology and ethics.
\newblock In \emph{Responsible Innovation 1}, pages 3--13. Springer
  Netherlands, Dordrecht.

\bibitem[{van Dijck(2013)}]{Van_Dijck2013-xf}
Jose van Dijck. 2013.
\newblock \emph{The culture of connectivity}.
\newblock Oxford University Press, New York, NY.

\bibitem[{van Dijck and Alinejad(2020)}]{Van_Dijck2020-rm}
Jos{\'e} van Dijck and Donya Alinejad. 2020.
\newblock Social media and trust in scientific expertise: Debating the covid-19
  pandemic in the netherlands.
\newblock \emph{Soc. Media Soc.}, 6(4):205630512098105.

\bibitem[{Vannini and Franzese(2008)}]{Vannini2008-ii}
Phillip Vannini and Alexis Franzese. 2008.
\newblock The authenticity of self: Conceptualization, personal experience, and
  practice.
\newblock \emph{Sociol. Compass}, 2(5):1621--1637.

\bibitem[{Vaswani et~al.(2017)Vaswani, Shazeer, Parmar, Uszkoreit, Jones,
  Gomez, Kaiser, and Polosukhin}]{Vaswani2017}
Ashish Vaswani, Noam Shazeer, Niki Parmar, Jakob Uszkoreit, Llion Jones,
  Aidan~N. Gomez, Lukasz Kaiser, and Illia Polosukhin. 2017.
\newblock \href
  {https://proceedings.neurips.cc/paper/2017/file/3f5ee243547dee91fbd053c1c4a845aa-Paper.pdf}
  {Attention is all you need}.
\newblock In \emph{Proceedings of the 31st International Conference on Neural
  Information Processing Systems}, pages 6000--6010, Long Beach, California,
  USA.

\bibitem[{Veale et~al.(2023)Veale, Matus, and Gorwa}]{veale2023ai}
Michael Veale, Kira Matus, and Robert Gorwa. 2023.
\newblock Ai and global governance: Modalities, rationales, tensions.
\newblock \emph{Annual Review of Law and Social Science}, 19.

\bibitem[{Verhoef et~al.(2021)Verhoef, Broekhuizen, Bart, Bhattacharya,
  Qi~Dong, Fabian, and Haenlein}]{Verhoef2021-cj}
Peter~C Verhoef, Thijs Broekhuizen, Yakov Bart, Abhi Bhattacharya, John
  Qi~Dong, Nicolai Fabian, and Michael Haenlein. 2021.
\newblock Digital transformation: A multidisciplinary reflection and research
  agenda.
\newblock \emph{J. Bus. Res.}, 122:889--901.

\bibitem[{von Schomberg(2013)}]{Von_Schomberg2013-hx}
Ren{\'e} von Schomberg. 2013.
\newblock A vision of responsible research and innovation.
\newblock In \emph{Responsible Innovation}, pages 51--74. John Wiley \& Sons,
  Ltd, Chichester, UK.

\bibitem[{Weidinger et~al.(2022)Weidinger, Uesato, Rauh, Griffin, Huang,
  Mellor, Glaese, Cheng, Balle, Kasirzadeh, Biles, Brown, Kenton, Hawkins,
  Stepleton, Birhane, Hendricks, Rimell, Isaac, Haas, Legassick, Irving, and
  Gabriel}]{weidinger}
Laura Weidinger, Jonathan Uesato, Maribeth Rauh, Conor Griffin, Po-Sen Huang,
  John Mellor, Amelia Glaese, Myra Cheng, Borja Balle, Atoosa Kasirzadeh,
  Courtney Biles, Sasha Brown, Zac Kenton, Will Hawkins, Tom Stepleton, Abeba
  Birhane, Lisa~Anne Hendricks, Laura Rimell, William Isaac, Julia Haas, Sean
  Legassick, Geoffrey Irving, and Iason Gabriel. 2022.
\newblock \href {https://doi.org/10.1145/3531146.3533088} {Taxonomy of risks
  posed by language models}.
\newblock In \emph{Proceedings of the 2022 ACM Conference on Fairness,
  Accountability, and Transparency}, FAccT '22, page 214–229, New York, NY,
  USA. Association for Computing Machinery.

\bibitem[{Wolfe et~al.(2023)Wolfe, Yang, Howe, and Caliskan}]{Wolfe}
Robert Wolfe, Yiwei Yang, Bill Howe, and Aylin Caliskan. 2023.
\newblock \href {https://doi.org/10.1145/3593013.3594072} {Contrastive
  language-vision ai models pretrained on web-scraped multimodal data exhibit
  sexual objectification bias}.
\newblock In \emph{Proceedings of the 2023 ACM Conference on Fairness,
  Accountability, and Transparency}, FAccT '23, page 1174–1185, New York, NY,
  USA. Association for Computing Machinery.

\bibitem[{Yarnold et~al.(2022)Yarnold, Maher, Hussey, and
  Dovers}]{Yarnold2022-ue}
Jennifer Yarnold, Ray Maher, Karen Hussey, and Stephen Dovers. 2022.
\newblock Uncertainty.
\newblock In \emph{Routledge Handbook of Global Environmental Politics}, pages
  253--268. Routledge, London.

\bibitem[{Zhang and Lu(2023)}]{zhang}
Annie~Li Zhang and Hang Lu. 2023.
\newblock \href {https://doi.org/10.1177/20563051231180623} {Scientists as
  influencers: The role of source identity, self-disclosure, and
  anti-intellectualism in science communication on social media}.
\newblock \emph{Social Media + Society}, 9(2):20563051231180623.

\bibitem[{Zhang et~al.(2018)Zhang, Wells, Wang, and Rohe}]{Zhang2018-nk}
Yini Zhang, Chris Wells, Song Wang, and Karl Rohe. 2018.
\newblock Attention and amplification in the hybrid media system: The
  composition and activity of donald trump's twitter following during the 2016
  presidential election.
\newblock \emph{New Media Soc.}, 20(9):3161--3182.

\end{thebibliography}
\bibliographystyle{acl_natbib}

\end{document}